\documentclass[11pt,a4paper]{article}
\usepackage[hyperref]{naaclhlt2019}
\usepackage{times}
\usepackage{latexsym}
\usepackage{breakurl}
\usepackage{url}

\usepackage{graphicx} 
\usepackage{amsmath} 
\usepackage{bm}
\usepackage{booktabs} 
\usepackage{multirow} 
\usepackage{color} 
\usepackage{subcaption} 

\aclfinalcopy 

\title{Learning to Describe Unknown Phrases with Local and Global Contexts}
\author{
{\bf Shonosuke Ishiwatari}$^{\dag}$ \quad
{\bf Hiroaki Hayashi}$^{\ddag}$ \quad
{\bf Naoki Yoshinaga}$^{\S}$ \quad
{\bf Graham Neubig}$^{\ddag}$ \quad \\
{\bf Shoetsu Sato}$^{\dag}$ \quad
{\bf Masashi Toyoda}$^{\S}$ \quad 
{\bf Masaru Kitsuregawa}$^{\P\S}$ \\
$\dag$ The University of Tokyo \quad
$\ddag$ Carnegie Mellon University\\
$\S$ Institute of Industrial Science, the University of Tokyo \quad
$\P$ National Institute of Informatics\\
    {$^{\dag\S\P}$\tt\{ishiwatari, ynaga, shoetsu, toyoda, kitsure\}@tkl.iis.u-tokyo.ac.jp}\\
{$^{\ddag}$\tt \{hiroakih, gneubig\}@cs.cmu.edu}
}

\date{}

\begin{document}
\maketitle
\begin{abstract}
When reading a text, it is common to become stuck on unfamiliar words and phrases, such as polysemous words with novel senses, rarely used idioms, internet slang, or emerging entities.
If we humans cannot figure out the meaning of those expressions from the immediate \emph{local} context, we consult dictionaries for definitions or search documents or the web to find other \emph{global} context to help in interpretation.
Can machines help us do this work?
Which type of context is more important for machines to solve the problem?
To answer these questions, we undertake a task of describing a given phrase in natural language based on its local and global contexts. To solve this task, we propose a neural description model that consists of two context encoders and a description decoder. In contrast to the existing methods for non-standard English explanation~\cite{ni2017learning} and definition generation~\cite{noraset2017definition,gadetsky2018conditional}, our model appropriately takes important clues from \emph{both} local and global contexts. Experimental results on three existing datasets (including WordNet, Oxford and Urban Dictionaries) and a dataset newly created from Wikipedia demonstrate the effectiveness of our method over previous work.
\end{abstract}

\section{Introduction} \label{sec:intro}
When we read news text with emerging entities, text in unfamiliar domains, or text in foreign languages, we often encounter expressions (words or phrases) whose senses we do not understand.
In such cases, we may first try to figure out the meanings of those expressions by reading the surrounding words (\emph{local} context) carefully.
Failing to do so, we may consult dictionaries, and in the case of polysemous words, choose an appropriate meaning based on the context.
Learning novel word senses via dictionary definitions is known to be more effective than contextual guessing~\cite{fraser:1998a,chen:2012a}.
However, very often, hand-crafted dictionaries do not contain definitions of expressions that are rarely used or newly created.
Ultimately, we may need to read through the entire document or even search the web to find other occurances of the expression (\emph{global} context) so that we can guess its meaning.

\begin{figure}[tb]
\centering
\includegraphics[width=1.0\linewidth]{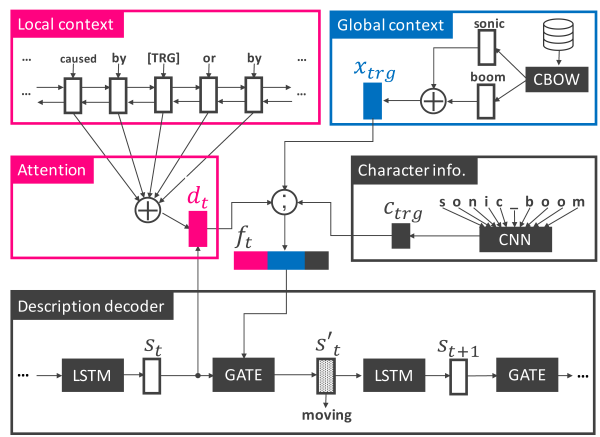}
\caption{\textbf{Lo}cal \& \textbf{G}lobal \textbf{C}ontext-\textbf{a}ware \textbf{D}escription generator (\textbf{LOG-CaD}).}
\label{fig/model}
\end{figure}

Can machines help us do this work?
\citet{ni2017learning} have proposed a task of generating a definition for a phrase given its local context.
However, they follow the strict assumption that the target phrase is newly emerged and there is only a single local context available for the phrase, which makes the task of generating an accurate and coherent definition difficult (perhaps as difficult as a human comprehending the phrase itself).
On the other hand, \citet{noraset2017definition} attempted to generate a definition of a word from an embedding induced from massive text (which can be seen as global context). This is followed by~\citet{gadetsky2018conditional} that refers to a local context to disambiguate polysemous words by choosing relevant dimensions of their word embeddings.
Although these research efforts revealed that both local and global contexts are useful in generating definitions, none of these studies exploited both contexts directly to describe unknown phrases.

In this study, we tackle the task of describing (defining) a phrase when given its local and global contexts.
We present \textbf{LOG-CaD}, a neural description generator (Figure~\ref{fig/model}) to directly solve this task.
Given an unknown phrase without sense definitions, our model obtains a phrase embedding as its global context by composing word embeddings while also encoding the local context.
The model therefore combines both pieces of information to generate a natural language description.

Considering various applications where we need definitions of expressions, we evaluated our method with four datasets including WordNet~\cite{noraset2017definition} for general words, the Oxford dictionary~\cite{gadetsky2018conditional} for polysemous words, Urban Dictionary~\cite{ni2017learning} for rare idioms or slang, and a newly-created Wikipedia dataset for entities. 

Our contributions are as follows:
\begin{itemize}
\item We propose \textbf{a general task of defining unknown phrases given their contexts}. This task is a generalization of three related tasks~\cite{noraset2017definition,ni2017learning,gadetsky2018conditional} and involves various situations where we need definitions of unknown phrases (\S~\ref{sec:task}).
\item We propose a method for \textbf{generating natural language descriptions for unknown phrases with local and global contexts} (\S~\ref{sec:proposal}).
\item As a benchmark to evaluate the ability of the models to describe entities, we \textbf{build a large-scale dataset} from Wikipedia and Wikidata for the proposed task. We release our dataset and the code\footnote{\url{https://github.com/shonosuke/ishiwatari-naacl2019}} to promote the reproducibility of the experiments (\S~\ref{sec:dataset}).
\item The proposed method achieves \textbf{the state-of-the-art performance} on our new dataset and the three existing datasets used in the related studies~\cite{noraset2017definition,ni2017learning,gadetsky2018conditional}  (\S~\ref{sec:experiments}).
\end{itemize}

\section{Context-aware Phrase Description Generation}
\label{sec:task}
In this section, we define our task of describing a phrase in a specific context.
Given an undefined phrase $X_{trg} = \{x_j, \cdots, x_k\}$ with its context $X = \{x_1, \cdots, x_I\}$ ($1 \leq j \leq k \leq I$), our task is to output a description $Y=\{y_1, \cdots, y_T\}$.
Here, $X_{trg}$ can be a word or a short phrase and is included in $X$.
$Y$ is a definition-like concrete and concise sentence that describes the $X_{trg}$.

For example, given a phrase ``sonic boom'' with its context ``the shock wave may be caused by \textit{sonic boom} or by explosion,'' the task is to generate a description such as ``sound created by an object moving fast.''
If the given context has been changed to ``this is the first official tour to support the band's latest studio effort, 2009's \textit{Sonic Boom},'' then the appropriate output would be ``album by Kiss.''

The process of description generation can be modeled with a conditional language model as
\begin{equation}
p(Y|X, X_{trg}) = \prod_{t=1}^T p(y_t | y_{<t}, X, X_{trg}). \label{eq/generation}
\end{equation}

\section{LOG-CaD: Local \& Global Context-aware Description Generator}
\label{sec:proposal}
In this section, we describe our idea of utilizing local and global contexts in the description generation task, and present the details of our model.

\subsection{Local \& global contexts}
When we find an unfamiliar phrase in text and it is not defined in dictionaries, how can we humans come up with its meaning?
As discussed in Section~\ref{sec:intro}, we may first try to figure out the meaning of the phrase from the immediate context, and then read through the entire document or search the web to understand implicit information behind the text.

In this paper, we refer to the explicit contextual information included in a given sentence with the target phrase (i.e., the $X$ in Eq.~\eqref{eq/generation}) as ``local context,'' and the implicit contextual information in massive text as ``global context.''
While both local and global contexts are crucial for humans to understand unfamiliar phrases, are they also useful for machines to generate descriptions?
To verify this idea, we propose to incorporate both local and global contexts to describe an unknown phrase.

\subsection{Proposed model}
Figure~\ref{fig/model} shows an illustration of our \textbf{LOG-CaD} model. 
Similarly to the standard encoder-decoder model with attention~\cite{bahdanau15,luong16}, it has a context encoder and a description decoder.
The challenge here is that the decoder needs to be conditioned not only on the local context, but also on its global context.
To incorporate the different types of contexts, we propose to use a gate function similar to~\citet{noraset2017definition} to dynamically control how the global and local contexts influence the description.

\paragraph{Local \& global context encoders}
We first describe how to model local and global contexts.
Given a sentence $X$ and a phrase $X_{trg}$, a bi-directional \textsc{lstm}~\cite{gers1999learning} encoder generates a sequence of continuous vectors $\bm{H} = \{\bm{h}_1 \cdots, \bm{h}_I\}$ as
\begin{equation}
\bm{h}_i = \mbox{Bi-LSTM} (\bm{h}_{i-1}, \bm{h}_{i+1}, \bm{x}_i),
\end{equation}
where $\bm{x}_i$ is the word embedding of word $x_i$.
In addition to the local context, we also utilize the global context obtained from massive text.
This can be achieved by feeding a phrase embedding $\bm{x}_{trg}$ to initialize the decoder~\cite{noraset2017definition} as
\begin{align}
    \bm{y}_0 &= \bm{x}_{trg}. \label{input}
\end{align}
Here, the phrase embedding $\bm{x}_{trg}$ is calculated by simply summing up all the embeddings of words that consistute the phrase $X_{trg}$.
Note that we use a randomly-initialized vector if no pre-trained embedding is available for the words in $X_{trg}$.

\paragraph{Description decoder}
Using the local and  global contexts, a description decoder computes the conditional probability of a description $Y$ with Eq.~\eqref{eq/generation}, which can be approximated with another \textsc{lstm} as
\begin{align}
\bm{s}_t &= \mbox{LSTM}(\bm{y}_{t-1}, \bm{s'}_{t-1}), \label{eq/lstm}\\
\bm{d}_t &= \mbox{ATTENTION}(\bm{H}, \bm{s}_t), \label{eq/attention}\\
\bm{c}_{trg} &= \mbox{CNN}(X_{trg}), \label{eq/cnn}\\
\bm{s}'_t &= \mbox{GATE}(\bm{s}_t, \bm{x}_{trg}, \bm{c}_{trg}, \bm{d}_t), \label{eq/gate}\\ 
p(y_t | y_{<t}, X_{trg}) &= \mbox{softmax}(\bm{W}_{s'} \bm{s}'_t + \bm{b}_{s'}) \label{eq/softmax},
\end{align}
where $\bm{s}_t$ is a hidden state of the decoder \textsc{lstm} ($\bm{s}_0 = \vec{0}$), and $\bm{y}_{t-1}$ is a jointly-trained word embedding of the previous output word $y_{t-1}$.
In what follows, we explain each equation in detail.

\paragraph{Attention on \textit{local} context}
Considering the fact that the local context can be relatively long (e.g., around 20 words on average in our Wikipedia dataset introduced in Section~\ref{sec:dataset}),
it is hard for the decoder to focus on important words in local contexts. 
In order to deal with this problem, the $\mbox{ATTENTION}(\cdot)$ function in Eq.~\eqref{eq/attention} decides which words in the local context $X$ to  focus on at each time step.
$\bm{d}_t$ is computed with an attention mechanism~\cite{luong16} as
\begin{align}
\bm{d}_t &= \sum_{i=1}^{T} \alpha_{i} \bm{h}_i, \\
\alpha_{i} &= \mbox{softmax}( \bm{U}_{h} \bm{h}_{i}^{\mathrm T} \bm{U}_{s} \bm{s}_t ),
\end{align}
where $\bm{U}_{h}$ and $\bm{U}_{s}$ are matrices that map the encoder and decoder hidden states into a common space, respectively.

\paragraph{Use of character information}
In order to capture the surface information of $X_{trg}$, we construct character-level \textsc{cnn}s~(Eq.~\eqref{eq/cnn}) following~\cite{noraset2017definition}.
Note that the input to the \textsc{cnn}s is a sequence of words in $X_{trg}$, which are concatenated with special character ``\_,'' such as ``sonic\_boom.''
Following \citeauthor{noraset2017definition} \shortcite{noraset2017definition}, we set the \textsc{cnn} kernels of length 2-6 and size ${10, 30, 40, 40, 40}$ respectively with a stride of 1 to obtain a 160-dimensional vector $\bm{c}_{trg}$.

\paragraph{Gate function to control local \& global contexts}
In order to capture the interaction between the local and global contexts, we adopt a $\mbox{GATE}(\cdot)$ function~(Eq.~\eqref{eq/gate}) which is similar to \citet{noraset2017definition}.
The $\mbox{GATE}(\cdot)$ function updates the LSTM output $\bm{s}_t$ to $\bm{s'}_t$ depending on the global context $\bm{x}_{trg}$, local context $\bm{d}_t$, and character-level information $\bm{c}_{trg}$ as
\begin{align}
    \bm{f}_t &= [\bm{x}_{trg} ; \bm{d}_t ; \bm{c}_{trg}] \label{feature}\\
    \bm{z}_t &= \sigma (\bm{W}_z [\bm{f}_t ; \bm{s}_t] + \bm{b}_z), \\
    \bm{r}_t &= \sigma (\bm{W}_r [\bm{f}_t ; \bm{s}_t] + \bm{b}_r), \\
    \bm{\tilde{s}}_t &= \mbox{tanh} (\bm{W}_s [(\bm{r}_t \odot \bm{f}_t) ; \bm{s}_t] + \bm{b}_s), \\
    \bm{s}'_t &= (1 - \bm{z}_t) \odot \bm{s}_t + z_t \odot \tilde{\bm{s}}_t,
\end{align}
where $\sigma(\cdot)$, $\odot$ and $;$ denote the sigmoid function, element-wise multiplication, and vector concatenation, respectively.
$\bm{W}_*$ and $\bm{b_*}$ are weight matrices and bias terms, respectively.
Here, the update gate $\bm{z}_t$ controls how much the original hidden state $\bm{s}_t$ is to be changed, and the reset gate $\bm{r}_t$ controls how much the information from $\bm{f}_t$ contributes to word generation at each time step.

\begin{figure*}[tb]
\centering
\includegraphics[width=1.0\linewidth]{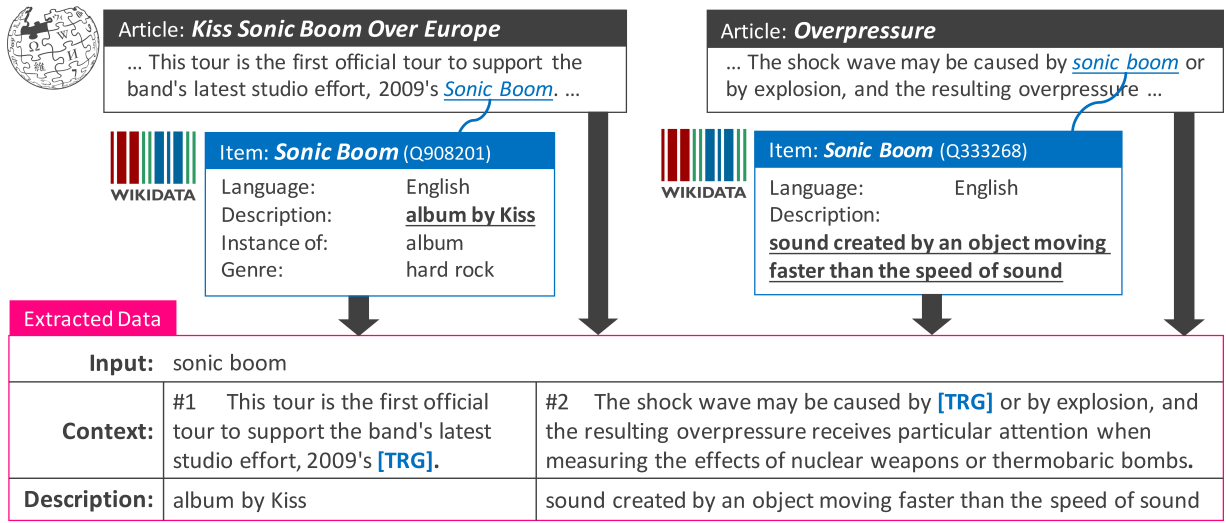}
\caption{Context-aware description dataset extracted from Wikipedia and Wikidata.}
\label{fig/dataset}
\end{figure*}

\section{Wikipedia Dataset}\label{sec:dataset}
Our goal is to let machines describe unfamiliar words and phrases, such as polysemous words, rarely used idioms, or emerging entities. Among the three existing datasets, WordNet and Oxford dictionary mainly target the words but not phrases, thus are not perfect test beds for this goal. On the other hand, although the Urban Dictionary dataset contains descriptions of rarely-used phrases, the domain of its targeted words and phrases is limited to Internet slang.

In order to confirm that our model can generate the description of
entities as well as polysemous words and slang,
we constructed a new dataset for context-aware phrase description generation from Wikipedia\footnote{\url{https://dumps.wikimedia.org/enwiki/20170720/}} and Wikidata\footnote{\url{https://dumps.wikimedia.org/wikidatawiki/entities/20170802/}} which contain a wide variety of entity descriptions with contexts.
The overview of the data extraction process is shown in Figure~\ref{fig/dataset}.
Each entry in the dataset consists of (1) a phrase, (2) its description, and (3) context (a sentence).

For preprocessing, we applied Stanford Tokenizer\footnote{\url{https://nlp.stanford.edu/software/tokenizer.shtml}}
to the descriptions of Wikidata items and the articles in Wikipedia. Next, we removed phrases in parentheses from the Wikipedia articles, since they tend to be paraphrasing in other languages and work as noise. 
To obtain the contexts of each item in Wikidata, we extracted the sentence which has a link referring to the item through all the first paragraphs of Wikipedia articles and replaced the phrase of the links with a special token \textbf{[TRG]}. 
Wikidata items with no description or no contexts are ignored.
This utilization of links makes it possible to resolve the ambiguity of words and phrases in a sentence without human annotations, which is a major advantage of using Wikipedia.
Note that we used only links whose anchor texts are identical to the title of the Wikipedia articles, since the users of Wikipedia sometimes link mentions to related articles.

\section{Experiments}
We evaluate our method by applying it to describe words in WordNet\footnote{\url{https://wordnet.princeton.edu/}}~\cite{miller1995wordnet} and Oxford Dictionary,\footnote{\url{https://en.oxforddictionaries.com/}} phrases in Urban Dictionary\footnote{\url{https://www.urbandictionary.com/}} and Wikipedia/Wikidata.\footnote{\url{https://www.wikidata.org}}
For all of these datasets, a given word or phrase has an inventory of senses with corresponding definitions and usage examples. These definitions are regarded as ground-truth descriptions.
\label{sec:experiments}

\paragraph{Datasets}
To evaluate our model on the word description task on WordNet, we followed~\citeauthor{noraset2017definition}~\shortcite{noraset2017definition} and extracted data from WordNet using the {\url{dict-definition}}\footnote{\url{https://github.com/NorThanapon/dict-definition}} toolkit.
Each entry in the data consists of three elements: (1) a word, (2) its definition, and (3) a usage example of the word.
We split this dataset to obtain Train, Validation, and Test sets.
If a word has multiple definitions/examples, we treat them as different entries.
Note that the words are mutually exclusive across the three sets.
The only difference between our dataset and theirs is that we extract the tuples only if the words have their usage examples in WordNet.
Since not all entries in WordNet have usage examples, our dataset is a small subset of \citet{noraset2017definition}.

\begin{table}[t]
\small
    \centering
        \begin{tabular}{@{}l@{\quad}r@{\quad}r@{\quad}r@{\quad}r@{\quad}r@{}}
            \toprule
            \textbf{Corpus} & \textbf{\#Phrases} & \textbf{\#Entries} & \textbf{Phrase} & \textbf{Context} & \textbf{Desc.}\\
            & & & {\bf length} & {\bf length} & {\bf length}\\
            \midrule
            \midrule
            \multicolumn{5}{@{}l}{WordNet} \\
            \midrule
            Train  & 7,938 & 13,883 & 1.00 & 5.81 & 6.61 \\
            Valid  & 998 & 1,752  & 1.00 & 5.64 & 6.61 \\
            Test   & 1,001 & 1,775  & 1.00 & 5.77 & 6.85 \\
            \midrule
            \midrule
            \multicolumn{5}{@{}l}{Oxford Dictionary} \\
            \midrule
            Train  & 33,128 & 97,855 & 1.00 & 17.74 & 11.02 \\
            Valid  & 8,867 & 12,232 & 1.00 & 17.80 & 10.99 \\
            Test   & 8,850 & 12,232 & 1.00 & 17.56 & 10.95 \\
            \midrule
            \midrule
            \multicolumn{5}{@{}l}{Urban Dictionary} \\
            \midrule
            Train  & 190,696 & 411,384 & 1.54 & 10.89 & 10.99 \\
            Valid  & 26,876 & 57,883 & 1.54 & 10.86 & 10.95 \\
            Test   & 26,875 & 38,371 & 1.68 & 11.14 & 11.50 \\
            \midrule
            \midrule
            \multicolumn{5}{@{}l}{Wikipedia} \\
            \midrule
            Train  & 151,995 & 887,455 & 2.10 & 18.79 & 5.89 \\
            Valid  & 8,361 & 44,003 & 2.11 & 19.21 & 6.31 \\
            Test   & 8,397  & 57,232 & 2.10 & 19.02 & 6.94 \\
            \bottomrule
        \end{tabular}
    \caption{Statistics of the word/phrase description datasets.\label{table/data}}
\end{table}

\begin{table}[t]
\small
    \centering
        \begin{tabular}{@{}lrrr@{}}
            \toprule
            {\bf Corpus} & {\bf Domain} & {\bf Inputs} & {\bf Cov. emb.} \\
            \midrule
            WordNet & General & words & $100.00\%$ \\
            Oxford Dictionary & General & words & $83.04\%$ \\
            Urban Dictionary & Internet slang & phrases & $21.00\%$ \\
            Wikipedia & Proper nouns & phrases & $26.79\%$ \\
            \bottomrule
        \end{tabular}
    \caption{Domains, expressions to be described, and the coverage of pre-trained embeddings of the expressions to be described.\label{table/properties}}
\end{table}

In addition to WordNet, we use the Oxford Dictionary following \citet{gadetsky2018conditional}, the Urban Dictionary following \citet{ni2017learning} and our Wikipedia dataset described in the previous section.
Table~\ref{table/data} and Table~\ref{table/properties} show the properties and statistics of the four datasets, respectively.

To simulate a situation in a real application where we might not have access to global context for the target phrases, we did not train domain-specific word embeddings on each dataset.
Instead, for all of the four datasets, we use the same pre-trained \textsc{cbow}\footnote{\texttt{GoogleNews-vectors-negative300.bin.gz} at \url{https://code.google.com/archive/p/word2vec/}} vectors trained on Google news corpus as global context following previous work~\cite{noraset2017definition,gadetsky2018conditional}.
If the expression to be described consists of multiple words, its phrase embedding is calculated by simply summing up all the \textsc{cbow} vectors of words in the phrase, such as ``sonic'' and ``boom.'' (See Figure~\ref{fig/model}).
If pre-trained \textsc{cbow} embeddings are unavailable, we instead 
use a special \textbf{[UNK]} vector (which is randomly initialized with a uniform distribution) as word embeddings.
Note that our pre-trained embeddings only cover $26.79\%$ of the words in the expressions to be described in our Wikipedia dataset, while it covers all words in WordNet dataset (See Table~\ref{table/properties}).
Even if no reliable word embeddings are available, all models can capture the character information through character-level \textsc{cnn}s (See Figure~\ref{fig/model}).

\paragraph{Models} 
\begin{table}[t]
\small
    \centering
        \begin{tabular}{@{}l@{\,\,\,}c@{\,\,\,}c@{\,\,\,}c@{\,\,\,}c@{}}
            \toprule
             & \textbf{Global} &  \textbf{Local} & \textbf{I-Attn.} & \textbf{LOG-CaD}\\
            \midrule
            \# Layers of Enc-LSTMs & - & 2  & 2 & 2 \\
            Dim. of Enc-LSTMs & - & 600 & 600 & 600 \\
            Dim. of Attn. vectors & - & 300 & 300 & 300 \\
            Dim. of input word emb. & 300 & - & 300 & 300 \\
            \hline
            Dim. of char. emb. & 160 & 160 & - & 160 \\
            \# Layers of Dec-LSTMs & 2 & 2 & 2 & 2 \\
            Dim. of Dec-LSTMs & 300 & 300 & 300 & 300 \\
            Vocabulary size & 10k & 10k & 10k & 10k \\
            Dropout rate & 0.5 & 0.5 & 0.5 & 0.5 \\
            \bottomrule
        \end{tabular}
    \caption{Hyperparameters of the models\label{table/parameter}}
\end{table}

We implemented four methods: (1) \textbf{Global}~\cite{noraset2017definition}, (2) \textbf{Local}~\cite{ni2017learning} with \textsc{cnn}, (3) \textbf{I-Attention}~\cite{gadetsky2018conditional}, and our proposed model, (4) \textbf{LOG-CaD}.
The \textbf{Global} model is our reimplementation of the best model (S + G + CH) in \citet{noraset2017definition}.
It can access the global context of a phrase to be described, but has no ability to read the local context.
The \textbf{Local} model is the reimplementation of the best model (dual encoder) in \citeauthor{ni2017learning}\shortcite{ni2017learning}.
In order to make a fair comparison of the effectiveness of local and global contexts, we slightly modify the original implementation by \citeauthor{ni2017learning}\shortcite{ni2017learning}; as the character-level encoder in the \textbf{Local} model, we adopt \textsc{cnn}s that are exactly the same as the other two models instead of the original \textsc{lstm}s.

The \textbf{I-Attention} is our reimplementation of the best model (S + I-Attention) in \citeauthor{gadetsky2018conditional}\shortcite{gadetsky2018conditional}.
Similar to our model, it uses both local and global contexts.
Unlike our model, however, it does not use character information to predict descriptions.
Also, it cannot directly use the local context to predict the words in descriptions.
This is because the \textbf{I-Attention} model indirectly uses the local context only to disambiguate the phrase embedding $\bm{x}_{trg}$ as
\begin{align}
\bm{x}'_{trg} &= \bm{x}_{trg} \odot \bm{m},\\
\bm{m} &= \sigma (\bm{W}_m  \frac{\sum_{i=1}^{I} \mbox{FFNN}(\bm{h}_i)}{I} + \bm{b}_m). \label{eq/ffnn}
\end{align}
Here, the $\mbox{FFNN}(\cdot)$ function is a feed-forward neural network that maps the encoded local contexts $\bm{h}_i$ to another space.
The mapped local contexts are then averaged over the length of the sentence $X$ to obtain a representation of the local context.
This is followed by a linear layer and a sigmoid function to obtain the soft binary mask $\bm{m}$ which can filter out the unrelated information included in global context.
Finally, the disambiguated phrase embedding $\bm{x}'_{trg}$ is then used to update the decoder hidden state as 
\begin{align}
\bm{s}_t &= \mbox{LSTM}([\bm{y}_{t-1} ; \bm{x}'_{trg}], \bm{s}_{t-1}).
\end{align}

All four models (Table~\ref{table/parameter}) 
are implemented with the PyTorch framework (Ver. 1.0.0).\footnote{\tt http://pytorch.org/}
\begin{table}[t]
 \small
    \centering
        \begin{tabular}{@{}lrrrr@{}}
            \toprule
            \textbf{Model} & WordNet & Oxford & Urban & Wikipedia \\
            \midrule
            \textbf{Global} & 24.10 & 15.05 & 6.05  & 44.77  \\
            \textbf{Local} & 22.34 & 17.90  & 9.03  & 52.94  \\
            \textbf{I-Attention} & 23.77 & 17.25 & 10.40  & 44.71  \\
            \textbf{LOG-CaD} & \textbf{24.79} & \textbf{18.53} & \textbf{10.55} & \textbf{53.85}  \\
            \bottomrule
        \end{tabular}
    \caption{\textsc{bleu} scores on four datasets.\label{table/bleu}}
\end{table}
\begin{table}[t]
 \small
    \centering
        \begin{tabular}{@{}lrr@{}}
            \toprule
            \textbf{Model} & \textbf{Annotated score}\\
            \midrule
            \textbf{Local} & 2.717 \\
            \textbf{LOG-CaD} & \textbf{3.008} \\
            \bottomrule
        \end{tabular}
    \caption{Averaged human annotated scores on Wikipedia dataset.\label{table/human}}
\end{table}
\paragraph{Automatic Evaluation}
Table~\ref{table/bleu} shows the \textsc{bleu}~\cite{papineni02} scores of the output descriptions.
We can see that the \textbf{LOG-CaD} model consistently outperforms the three baselines in all four datasets.
This result indicates that using both local and global contexts helps describe the unknown words/phrases correctly.
While the \textbf{I-Attention} model also uses local and global contexts, its performance was always lower than the \textbf{LOG-CaD} model.
This result shows that using local context to predict description is more effective than using it to disambiguate the meanings in global context.

In particular, the low \textsc{bleu} scores of \textbf{Global} and \textbf{I-Attention} models on 
Wikipedia dataset suggest that it is necessary to learn to ignore the noisy information in global context if the coverage of pre-trained word embeddings is extremely low (see the third and fourth rows in Table~\ref{table/properties}).
We suspect that the Urban Dictionary task is too difficult and the results are unreliable considering its extremely low \textsc{bleu} scores and high ratio of unknown tokens in generated descriptions.

\begin{table}[t]
\footnotesize
    \centering
        \begin{tabular}{@{}r@{\,\,\,\,}l@{\quad}l@{}}
            \toprule
            \textbf{Input:} & waste & \\
            \midrule
            \multirow{2}{*}{\textbf{Context:}} & \textbf{$\#1$} & \textbf{$\#2$}\\
            \cmidrule{2-3}
            & \multicolumn{1}{@{}p{2.8cm}}{if the effort brings no compensating gain it is a \textbf{waste}} & \multicolumn{1}{@{}p{2.8cm}@{}}{We \textbf{waste} the dirty water by channeling it into the sewer} \\
            \midrule
            \textbf{Reference:} & \multicolumn{1}{@{}p{2.8cm}}{useless or profitless activity} & to get rid of \\ \midrule
            \textbf{Global:} & \multicolumn{2}{@{}l}{to give a liquid for a liquid} \\ \midrule
            \textbf{Local:} & \multicolumn{1}{@{}p{2.8cm}}{a state of being assigned to a particular purpose} & \multicolumn{1}{@{}p{2.8cm}@{}}{to make a break of a wooden instrument} \\ \midrule
            \textbf{I-Attention:} & \multicolumn{1}{@{}p{2.8cm}}{a person who makes something that can be be be done} & \multicolumn{1}{@{}p{2.8cm}@{}}{to remove or remove the contents of} \\ \midrule
            \textbf{LOG-CaD:} & \multicolumn{1}{@{}p{2.8cm}@{}}{a source of something that is done or done}
 &to remove a liquid \\
            \bottomrule
        \end{tabular}
    \caption{Descriptions for a word in WordNet. \label{table/examples_wordnet2}}
\end{table}

\begin{table}[t]
\footnotesize
    \centering
        \begin{tabular}{@{}r@{\,\,\,\,}l@{\quad}l@{}}
            \toprule
            \textbf{Input:} & daniel o'neill & \\
            \midrule
            \multirow{2}{*}{\textbf{Context:}} & \textbf{$\#1$} & \textbf{$\#2$}\\
            \cmidrule{2-3}
            & \multicolumn{1}{@{}p{2.8cm}}{after being enlarged by publisher \textbf{daniel o'neill} it was reportedly one of the largest and most prosperous newspapers in the united states.} & \multicolumn{1}{@{}p{2.8cm}@{}}{in 1967 he returned to belfast where he met fellow belfast artist \textbf{daniel o'neill}.} \\
            \midrule
            \textbf{Reference:} & american journalist & irish artist\\ \midrule
            \textbf{Global:} & american musician  \\
            \midrule
            \textbf{Local:} & american publisher & british musician\\ \midrule
            \textbf{I-Attention:} & american musician & american musician\\ \midrule
            \textbf{LOG-CaD:} & american writer & british musician\\
            \midrule
        \end{tabular}
\caption{Descriptions for a phrase in Wikipedia. \label{table/examples_wiki1}}
\end{table}

\begin{table*}
        \footnotesize
    \centering
        \begin{tabular}{@{}rllll@{}}
            \midrule
            \textbf{Input:} & q & & &\\
            \midrule        
            \multirow{2}{*}{\textbf{Context:}} & \textbf{$\#1$} & \textbf{$\#2$} & \textbf{$\#3$}& \textbf{$\#4$} \\
            \cmidrule{2-5}
            & \multicolumn{1}{p{3cm}}{q-lets and co. is a filipino and english informative children 's show on \textbf{q} in the philippines .} & \multicolumn{1}{p{3cm}}{she was a founding producer of the cbc radio one show '' \textbf{q} '' .} & \multicolumn{1}{p{3cm}}{the q awards are the uk 's annual music awards run by the music magazine '' \textbf{q} '' .} & \multicolumn{1}{p{3.5cm}}{charles fraser-smith was an author and one-time missionary who is widely credited as being the inspiration for ian fleming 's james bond quartermaster \textbf{q} .}\\
            \midrule
            \textbf{Reference:} & philippine tv network & canadian radio show & british music magazine & \multicolumn{1}{p{3cm}}{fictional character from james bond}\\ \midrule
            \textbf{Global:} & american rapper \\
            \midrule
            \textbf{Local:} & television channel & television show show & magazine  & american writer\\ \midrule
            \textbf{I-Attention:} & american rapper & american rapper  & american rapper & american rapper \\ \midrule
            \textbf{LOG-CaD:} & \multicolumn{1}{p{3cm}}{television station in the philippines} & television program & \multicolumn{1}{p{3cm}}{british weekly music journalism magazine} & [unk] [unk]\\
            \bottomrule
        \end{tabular}
    \caption{Descriptions for a word in Wikipedia. \label{table/examples_wiki2}}
\end{table*}

\paragraph{Manual Evaluation}
To compare the proposed model and the strongest baseline in Table~\ref{table/bleu} (i.e., the \textbf{Local} model), we performed a human evaluation on our dataset.
We randomly selected 100 samples from the test set of the Wikipedia dataset and asked three native English speakers to rate the output descriptions from 1 to 5 points as: 1) completely wrong or self-definition, 2) correct topic with wrong information, 3) correct but incomplete, 4) small details missing, 5) correct.
The averaged scores are reported in Table~\ref{table/human}.
Pair-wise bootstrap resampling test~\cite{koehn:2004:EMNLP} for the annotated scores has shown that the superiority of \textbf{LOG-CaD} over the \textbf{Local} model is statistically significant ($p < 0.01$).

\paragraph{Qualitative Analysis}
Table~\ref{table/examples_wordnet2} shows a word in the WordNet, while Table~\ref{table/examples_wiki1} and Table~\ref{table/examples_wiki2} show the examples of the entities in Wikipedia as examples.
When comparing the two datasets, the quality of generated descriptions of Wikipedia dataset is significantly better than that of WordNet dataset.
The main reason for this result is 
that the size of training data of the Wikipedia dataset is 64x larger than the WordNet dataset (See Table~\ref{table/data}).

For all examples in the three tables, the \textbf{Global} model can only generate a single description for each input word/phrase because it cannot access any local context.
In the Wordnet dataset, only the \textbf{I-Attention} and \textbf{LOG-CaD} models can successfully generate the concept of ``remove'' given the context $\#2$.
This result suggests that considering both local and global contexts are essential to generate correct descriptions.
In our Wikipedia dataset, both the \textbf{Local} and \textbf{LOG-CaD} models can describe the word/phrase considering its local context.
For example, both the \textbf{Local} and \textbf{LOG-CaD} models could generate ``american'' in the description for ``daniel o'neill'' given ``united states'' in context $\#1$, while they could generate ``british'' given ``belfast'' in context $\#2$.
A similar trend can also be observed in Table~\ref{table/examples_wiki2}, where \textbf{LOG-CaD} could generate the locational expressions such as ``philippines'' and ``british'' given the different contexts.
On the other hand, the \textbf{I-Attention} model could not describe the two phrases, taking into account the local contexts.
We will present an analysis of this phenomenon in the next section.

\begin{figure*}[tb]
\centering
  \begin{subfigure}[b]{0.31\textwidth} 
    \centering
    \includegraphics[width=1.0\linewidth]{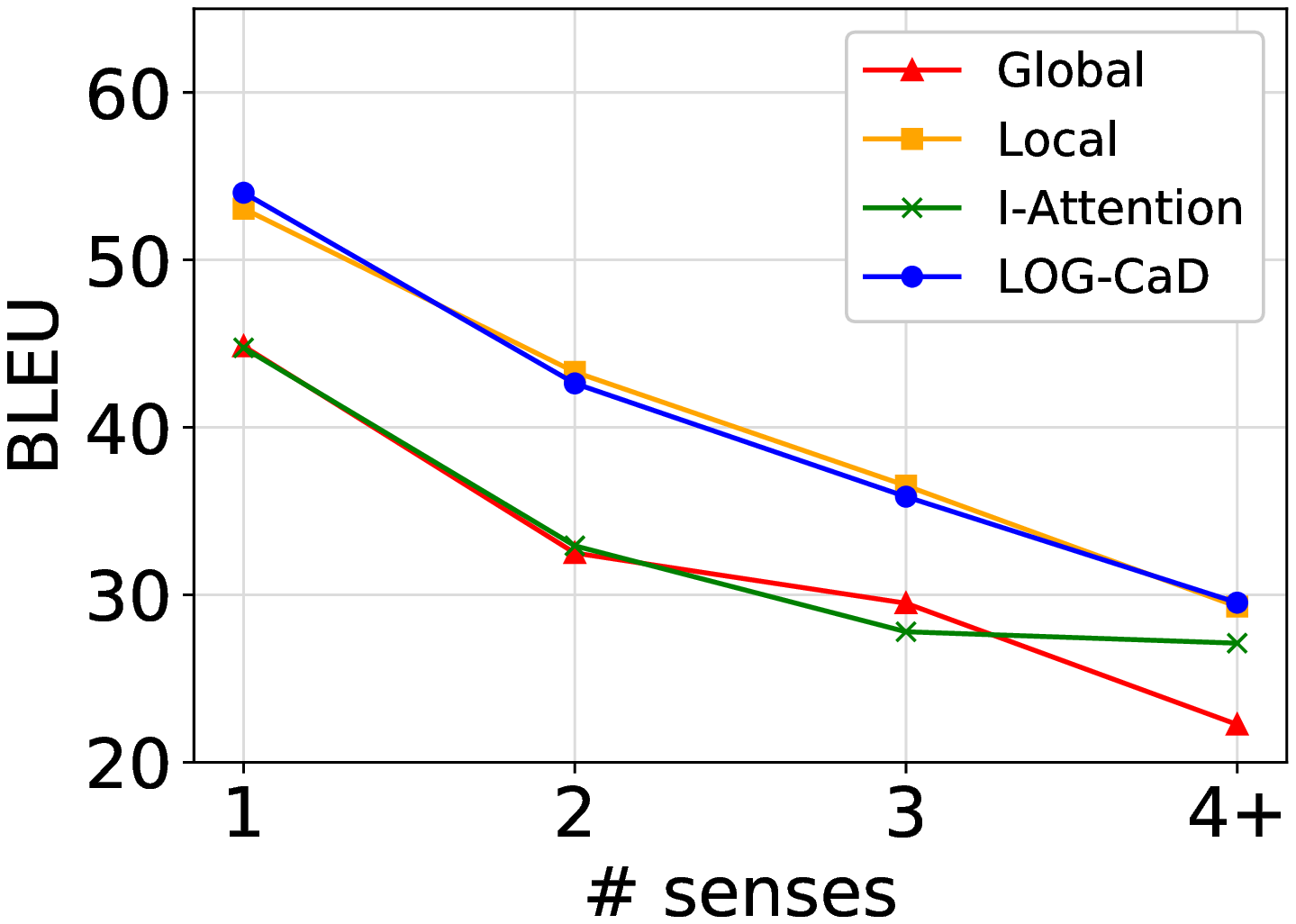}
    \caption{Number of senses of the phrase.}
    \label{fig/num_sense}
  \end{subfigure}\quad%
  \begin{subfigure}[b]{0.31\textwidth}
    \centering
    \includegraphics[width=1.0\linewidth]{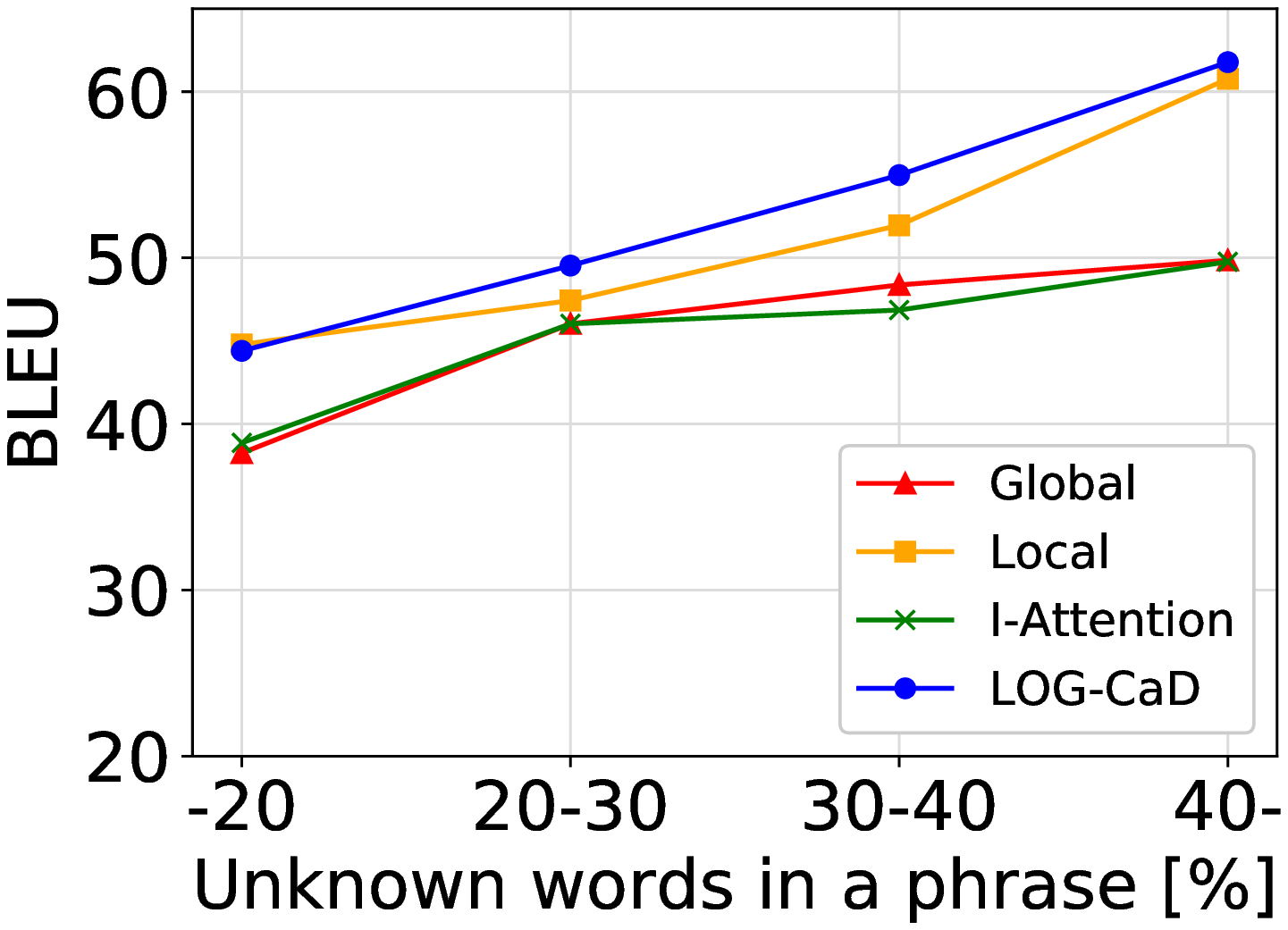}
    \caption{Unknown words ratio in the phrase.}
    \label{fig/ratio_unk}
  \end{subfigure}\quad%
  \begin{subfigure}[b]{0.31\textwidth}
    \centering
    \includegraphics[width=1.0\linewidth]{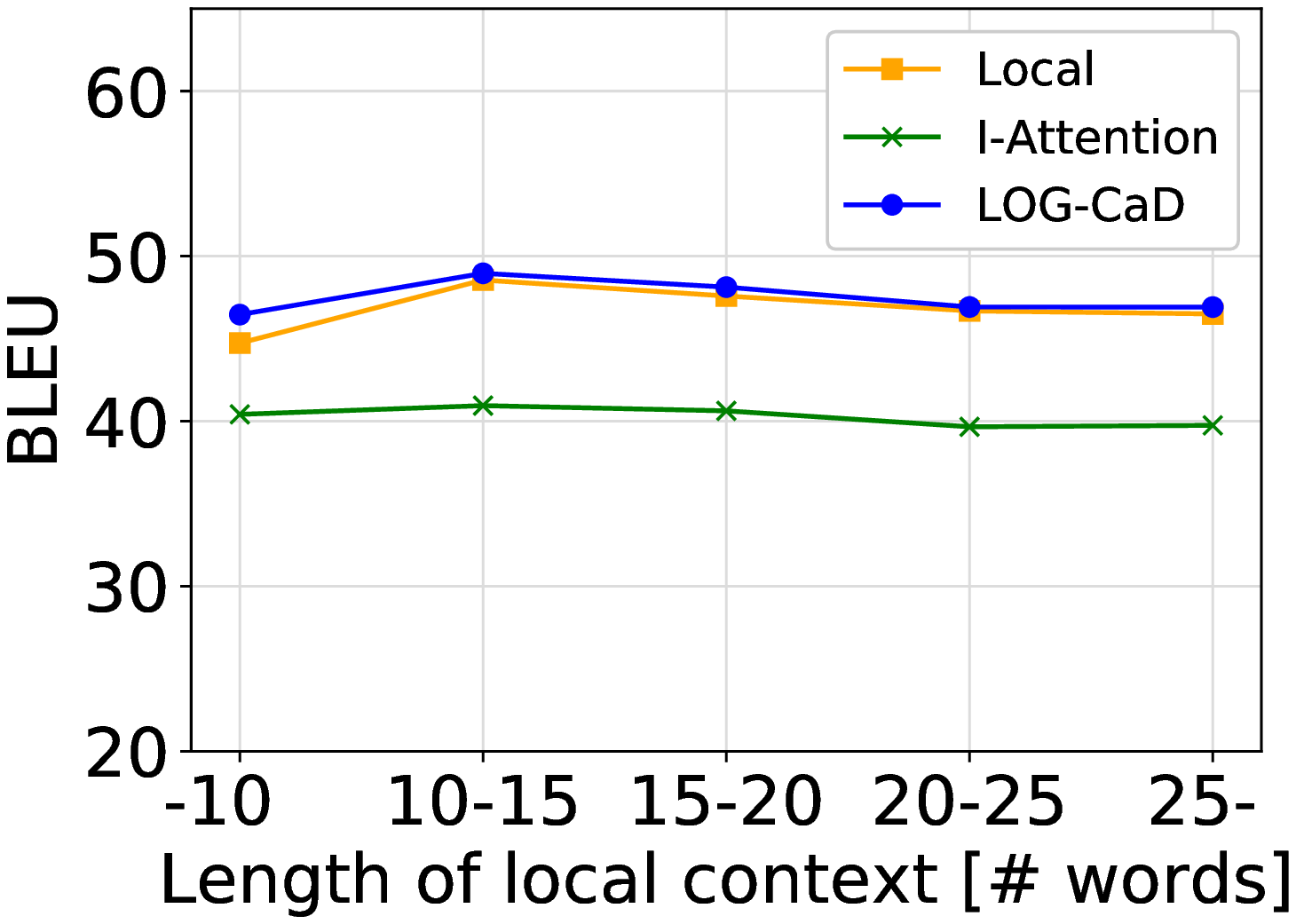}
    \caption{Length of the local context.}
  \end{subfigure}
  \caption{Impact of various parameters of a phrase to be described on \textsc{bleu} scores of the generated descriptions. \label{fig/sense_unk_len}}
\end{figure*}

\section{Discussion}
In this section, we present analyses on how the local and global contexts contribute to the description generation task.
First, we discuss how the local context helps the models to describe a phrase.
Then, we analyze the 
impact
of global context under the situation where local context is unreliable.

\subsection{How do the models utilize \textit{local} contexts?}
Local context helps us (1) disambiguate polysemous words and (2) infer the meanings of unknown expressions.
Can machines also utilize the local context?
In this section, we discuss the two roles of local context in description generation.

Considering that the pre-trained word embeddings are obtained from word-level co-occurrences in massive text, more information is mixed up into a single vector as the more senses the word has.
While \citeauthor{gadetsky2018conditional}~\shortcite{gadetsky2018conditional} designed the \textbf{I-Attention} model to filter out unrelated meanings in the global context given local context, they did not discuss the impact of the number of senses has on the performance of definition generation.
To understand the influence of the ambiguity of phrases to be defined on the generation performance, we did an analysis on our Wikipedia dataset. Figure~\ref{fig/sense_unk_len}(a) 
shows that the description generation task becomes harder as the phrases to be described become more ambiguous.
In particular, when a phrase has an extremely large number of senses, (i.e., \#senses $\geq 4$), the \textbf{Global} model drops its performance significantly.
This result indicates that the local context is necessary to disambiguate the meanings in global context.

As shown in Table~\ref{table/properties}, a large proportion of the phrases in our Wikipedia dataset includes unknown words (i.e., only $26.79\%$ of words in the phrases have their pre-trained embeddings).
This fact indicates that the global context in this dataset is not fully reliable.
Then our next question is, how does the lack of information from global context affect the performance of phrase description?
Figure~\ref{fig/sense_unk_len}(b) shows the impact of unknown words in the phrases to be described on the performance.
As we can see from the result, the advantage of \textbf{LOG-CaD} and \textbf{Local} models over \textbf{Global} and \textbf{I-Attention} models becomes larger as the unknown words increases.
This result suggests that we need to fully utilize local contexts especially in practical applications where the phrases to be defined have many unknown words.
Here, Figure~\ref{fig/sense_unk_len}(b) also shows a counterintuitive phenomenon that \textsc{bleu} scores increase as the ratio of unknown words in a phrase increase. This is mainly because unknown phrases tend to be person names such as writers, actors, or movie directors. Since these entities have fewer ambiguities in categories, they can be described in extremely short sentences that are easy for all four models to decode (e.g., ``finnish writer'' or ``american television producer'').

\subsection{How do the models utilize \textit{global} contexts?}
As discussed earlier, local contexts are important to describe unknown expressions, but how about global contexts?
Assuming a situation where we cannot obtain much information from local contexts (\textit{e.g.}, infer the meaning of ``boswellia'' from a short local context ``Here is a boswellia''), global contexts should be essential to understand the meaning.
To confirm this hypothesis, we analyzed the impact of the length of local contexts on \textsc{bleu} scores. Figure~\ref{fig/sense_unk_len}(c) shows that when the length of local context is extremely short ($l \leq 10$), the \textbf{LOG-CaD} model becomes much stronger than the \textbf{Local} model.
This result indicates that not only local context but also global context help models describe the meanings of phrases.

\section{Related Work}
\label{sec:related}
In this study, we address a task of describing a given phrase with its context. In what follows, we explain existing tasks that are related to our work.

Our task is closely related to word sense disambiguation (\textsc{wsd})~\cite{Navigli:2009a}, which identifies a pre-defined sense for the target word with its context. 
Although we can use it to solve our task by retrieving the definition sentence for the sense identified by \textsc{wsd}, it requires a substantial amount of training data to handle a different set of meanings of each word, and cannot handle words (or senses) which are not registered in the dictionary.
Although some studies have attempted to detect novel senses of words for given contexts~\cite{erk:2006a,lau:2014a}, they do not provide definition sentences.
Our task avoids these difficulties in \textsc{wsd} by directly generating descriptions for phrases or words. 
It also allows us to flexibly tailor a fine-grained definition for the specific context.

Paraphrasing~\cite{androutsopoulos:2010a,madnani:2010a} (or text simplification~\cite{siddharthan:2014a}) can be used to rephrase words with
unknown senses. 
However, the target of paraphrase acquisition are words/phrases with no specified context.
Although a few studies \cite{connor:2007a,max:2009a,max:2012a} consider sub-sentential (context-sensitive) paraphrases, they do not intend to obtain a definition-like description as a paraphrase of a word.

Recently, \citeauthor{noraset2017definition}~\shortcite{noraset2017definition} introduced a task of generating a definition sentence of a word from its pre-trained embedding.
Since their task does not take local contexts of words as inputs, their method cannot generate an appropriate definition for a polysemous word for a specific context.
To cope with this problem, \citeauthor{gadetsky2018conditional} \shortcite{gadetsky2018conditional} proposed a definition generation method that works with polysemous words in dictionaries.
They presented a model that utilizes local context to filter out the unrelated meanings from a pre-trained word embedding in a specific context.
While their method use local context for disambiguating the meanings that are mixed up in word embeddings, the information from local contexts cannot be utilized if the pre-trained embeddings are unavailable or unreliable.
On the other hand, our method can fully utilize the local context through an attentional mechanism, even if the reliable word embeddings are unavailable.

The most related work to this paper is \citet{ni2017learning}.
Focusing on non-standard English phrases, they proposed a model to generate the explanations solely from local context.
They followed the strict assumption that the target phrase was newly emerged and there was only a single local context available, which made the task of generating an accurate and coherent definition difficult.
Our proposed task and model are more general and practical than \citet{ni2017learning}; where (1) we use Wikipedia, which includes expressions from various domains, and (2) our model takes advantage of global contexts if available.

Our task of describing phrases with its context is a generalization of the three tasks~\cite{noraset2017definition,ni2017learning,gadetsky2018conditional}, and the proposed method utilizes both local and global contexts of an expression in question.

\section{Conclusions}
This paper sets up a task of generating a natural language description for an unknown phrase with a specific context, aiming to help us acquire unknown word senses when reading text. 
We approached this task by using a variant of encoder-decoder models that capture the given local context with the encoder and global contexts with the decoder initialized by the target phrase's embedding induced from massive text.
We performed experiments on three existing datasets and one newly built from Wikipedia and Wikidata.
The experimental results confirmed that the local and global contexts complement one another and are both essential; global contexts are crucial when local contexts are short and vague, while the local context is important when the target phrase is polysemous, rare, or unseen.

As future work, we plan to modify our model to use multiple contexts in text to improve the quality of descriptions, considering the ``one sense per discourse'' hypothesis~\cite{gale:1992a}.
We will release the newly built Wikipedia dataset and the experimental codes for the academic and industrial communities at \url{https://github.com/shonosuke/ishiwatari-naacl2019} to facilitate the reproducibility of our results and their use in various application contexts.

\section*{Acknowledgements}
The authors are grateful to Thanapon Noraset for sharing the details of his implementation of the previous work.
We also thank the anonymous reviewers for their careful reading of our paper and insightful comments, and the members of Kitsuregawa-Toyoda-Nemoto-Yoshinaga-Goda laboratory in the University of Tokyo for proofreading the draft.

This work was partially supported by Grant-in-Aid for JSPS Fellows (Grant Number 17J06394) and Commissioned Research (201) of the National Institute of Information and Communications Technology of Japan.

\bibliography{naaclhlt2019}
\bibliographystyle{acl_natbib}
\end{document}